\documentclass[conference]{IEEEtran}

\usepackage{xcolor}
\usepackage{algorithm}
\usepackage{algpseudocode}
\usepackage{bm}
\usepackage{tabularx}
\usepackage{multirow}
\usepackage{rotating}
\usepackage{array}
\usepackage{booktabs}
\usepackage{amsmath}
\usepackage{caption}
\usepackage{soul}

\newenvironment{changemargin}[2]{\list{}{\rightmargin#2\leftmargin#1\parsep0pt\topsep0pt\partopsep0pt}\item[]}{\endlist}

\title{FL-NAS: Towards Fairness of NAS for Resource Constrained Devices via Large Language Models
}

\author{
\IEEEauthorblockN{\textbf{\large{(Invited Paper)}}}
\IEEEauthorblockN{
Ruiyang Qin\textsuperscript{\textdagger}, 
Yuting Hu\textsuperscript{\textsection}, 
Zheyu Yan\textsuperscript{\textdagger}, 
Jinjun Xiong\textsuperscript{\textsection}, 
Ahmed Abbasi\textsuperscript{\textdagger}, 
Yiyu Shi\textsuperscript{\textdagger}
}
\IEEEauthorblockA{
\textsuperscript{\textdagger}University of Notre Dame, 
\textsuperscript{\textsection}University at Buffalo
}
\IEEEauthorblockA{
(For correspondence, contact jinjun@buffalo.edu, yshi4@nd.edu)
}
}

\begin{document}

\maketitle

\begin{abstract}
    Neural Architecture Search (NAS) has become the de fecto tools in the industry in automating the design
    of deep neural networks for various applications, especially those driven by mobile and edge devices
    with limited computing resources. The emerging large language models (LLMs), due to their prowess,  have also 
    been incorporated into NAS recently and show some promising results. This paper conducts
     further exploration in this direction by considering three important design metrics simultaneously, i.e.,
     model accuracy, fairness, and hardware deployment efficiency. We propose a novel LLM-based NAS
     framework, FL-NAS, in this paper, and show experimentally that FL-NAS can indeed find high-performing
     DNNs, beating state-of-the-art DNN models by orders-of-magnitude across almost all design considerations. 
    \end{abstract}
    

\section{Introduction}
\label{sec_intro}

\noindent
Deep Neural Networks (DNNs) have emerged as a powerful machine learning model in modern artificial intelligence (A.I.) research.
In many applications,
DNNs have demonstrated superior performance, sometimes even surpassing human's performance. To alleviate the manual and expertise-driven 
design efforts for DNNs, 
Neural Architecture Search (NAS)~\cite{elsken2019neural} has made significant progresses recently in automating the design of DNNs. 
It is not exaggerating to say that NAS has become the cornerstone in democratization of DNNs for all. However, a major drawback of
NAS is its exorbitant computational cost, which made it less accessible for the general public, hence biasing many of the A.I. research outcomes to
only those who can afford to having large computing servers with high-performance graphic processing units (GPUs). 
This raises one aspect of the fairness issue in A.I. research. 

Another equally important aspect of fairness is DNNs' 
evaluation metrics. Traditionally, DNNs are evaluated for  accuracy for a given task. 
But given the wide applications of DNN models to various societally critical applications, fairness of DNN models' prediction across
different demographic groups (such as gender, age, races etc.) as presented in the dataset has risen as another important research topic.
Hence in term, there are some recent NAS efforts to incorporate the fairness considerations~\cite{chu2021fairnas, liu2021fnas, sheng2022larger}.

The wide applications of DNN models have unavoidably met with another deployment challenge, i.e., the deployment of DNNs to
increasingly popular mobile and edge computing devices, such as healthcare devices~\cite{jia2023importance}
and medical diagnostic systems~\cite{kaissis2020secure}, which possess limited computing resources (such as memory sizes and power). 
Various techniques have been proposed address this challenge, such as 
model compression, pruning, quantization, and software-hardware co-design~\cite{han2015deep, yan2022reliability, yan2022swim, nagel2021white}. 
Many of those techniques have further been incorporated into recent NAS work. But most of them are still focused on optimizing the accuracy-related metrics without 
considering the fairness metrics. One exception is the recent work of FaHaNa~\cite{sheng2022larger}.

A very recent trend in A.I. is the foundational models as represented by the large 
language models (LLMs)~\cite{OpenAI2023GPT4TR, touvron2023llama, biderman2023pythia, anil2023palm}. Trained on massive amounts of data
with trillions of parameters, LLMs have garnered the well-deserving limelight for their exceptional performance
on a wide range of tasks, demonstrating deep knowledge and reasoning capabilities across a wide range of domains. 
Because of LLMs' seemingly universal prowess, researchers have started to explore the application of LLMs
for NAS~\cite{yan2023viability, zheng2023can, wang2023graph, li2023diffnas}. Though some interesting results have
been reported, they exhibited the same drawback as traditional NAS solely focusing on accuracy metrics, without considering fairness metrics.

Therefore, in this paper, we propose to re-examine the NAS problem under LLMs while considering three related design metrics simultaneously, i.e., accuracy,
fairness, and hardware efficiency (i.e., hardware devices under limited computing resources). 
We introduce a novel NAS approach, FL-NAS (\textbf{F}airness with \textbf{L}arge Language Models on \textbf{N}eural \textbf{A}rchitecture \textbf{S}earch), to leverage LLMs' capabilities 
in understanding the semantic subtleties of much harder design considerations, i.e., hardware efficiency and fairness. 
We empirically show that FL-NAS can reasonably
incorporate both fairness and hardware efficiency in its design consideration through LLMs, yet without sacrificing accuracy.
In other words, through proper prompt designs,
LLMs can indeed suggest improved DNN architectures in terms of model accuracy, fairness, and hardware efficiency. For example, compared
to state-of-the-art (SOTA) work under the same experimental settings, FL-NAS can improve the model accuracy by up to 11.4\% and fairness scores by up to 18$\times$, respectively, and at the same time, reduce the model
inference latency by up to 4$\times$, the DNN model
size by up to 18$\times$, and memory footprint by up to 55$\times$, respectively.
This opens up new opportunities for designing better prompts for LLMs to guide the NAS for better DNNs while considering 
some of the traditionally more subtle design considerations than before.

The rest of the paper is structured as follows. Section~\ref{sec_background} introduces the background and motivates our work;
Section~\ref{sec_method} discusses the details about FL-NAS design;
Section~\ref{sec_exp} shows the experimental results and analyses;
and finally Section~\ref{sec_diss} concludes the paper with discussion of our future work.


\section{Background and Motivations}
\label{sec_background}

\subsection{Neural Architecture Search}
\noindent
Neural Architecture Search (NAS)~\cite{zoph2016neural} is a technique to automatically design a deep neural network (DNN) architecture. It does this by exploring
a large design space of candidate DNN architectures while meeting certain design objectives, where
 model accuracy has been the primary concern traditionally. 
Evolutionary algorithms (EAs)~\cite{liu2021survey}  and reinforcement learning (RL)~\cite{baker2017accelerating} are two of
the mostly utilized methods for NAS. For example,  RL-based NAS methods utilize various RL algorithms (e.g.,
deep Q-learning and policy gradient methods) to effectively guide the search process towards better architectures~\cite{baker2017accelerating}.
EA-based NAS techniques~\cite{liu2021survey} utilize principles from evolutionary computation by
employing mutation, recombination, and selection to evolve neural architectures. Both approaches 
have seen significant progresses in recent years, and the essence of those approaches is to effectively
balance the need of exploring intricate and large  design spaces and the cost of high computation. 

\subsection{NAS under Resource and Fairness Constraints}
\noindent
As mobile and edge computing increasingly becomes the dominant deployment 
platform for DNN-based applications, NAS has been extended to consider more practical
design objectives beyond accuracy. The first of such consideration is the hardware efficiency, i.e.,
how to design DNNs that can be effectively deployed onto resource-constrained 
devices~\cite{zhang2019neural, stamoulis2019single, li2020edd}.
The second consideration is fairness, which has garnered increasing attention recently.
Different from accuracy and hardware efficiency, which are relatively easy to quantify, fairness
delves into the nuanced output differences among various demographic groups within the data, such as
genders and ages. FaHaNa~\cite{sheng2022larger} is one of the few works\footnote{Other interpretations
of fairness that are not related to demographics are possible, but they is outside of this paper's scope, such as~\cite{chu2021fairnas, liu2021fnas}}
that incorporated both   fairness and hardware-efficiency into its NAS search along with the conventional accuracy metrics.

\subsection{NAS via Large Language Models}
\noindent
Since there are two major contributors to the high cost for NAS, i.e., the evaluation of candidate architectures' metrics
and the search of new candidate architecture, 
the most recent development of NAS has seen two interesting solutions.
The first one is to develop proxy-based NAS methods~\cite{cai2018proxylessnas, zhou2020econas, Li_2023_ICCV} where the evaluation of DNN architectures
is replaced with a more efficient surrogate proxy metric, hence speeding up the costly NAS search process.
The second one is to leverage the latest development of large language models (LLMs) that
has shown powerful semantic
understanding capabilities across natural languages, source codes, and various application 
domains. In other words, NAS search can be replaced with prompting engineering for
LLMs so that it would suggest new DNN architecture candidates with the desired quality improvement.
Since LLMs' inference time is orders of magnitude faster
than NAS search time, LLM-based NAS also promises to be faster than conventional NAS. 

LCDA~\cite{yan2023viability} may be one of the first work to explore the potential of utilizing LLMs in deploying computing-in-memory based DNNs on edge devices. Other works include
GENIUS~\cite{zheng2023can} that utilized LLMs in NAS to improve the accuracy of convolution-based DNNs,  GPT4GNAS~\cite{wang2023graph} that explores the potential of LLMs in improving the accuracy of graph neural networks, and DIFFNAS~\cite{li2023diffnas} that uses LLMs to search for DNNs that are more compatible with Denoising Diffusion Probabilistic Models (DDPM).
All these work share the same patterns as to design proper prompts for LLMs so that
LLMs can focus on the improvement of design metrics for the given DNN architecture templates. 
Though they have shown a promising path to use LLMs to speed up NAS, none has
considered the fairness as part of their design metrics.

Given LLMs' strength in understanding the ``vagueness'' of natural languages, it
seems logical to use LLMs to interpret the meaning of fairness, hence allowing
LLMs to suggest better DNN candidates that can also exhibit good fairness scores.
This forms the basis of our work in this paper. In other words,
we would like to use LLMs to help NAS evaluate the goodness of a DNN candidate
in terms of all the desired design objectives, including accuracy, hardware efficiency,
and fairness. To do so, we propose a novel FL-NAS framework as discussed in the next section.



\section{FL-NAS Design Details}
\label{sec_method}

\noindent
We propose FL-NAS, \textbf{F}airness with \textbf{L}arge Language Models on \textbf{N}eural \textbf{A}rchitecture \textbf{S}earch,
 as a new NAS framework that utilizes LLMs to efficiently search for DNN architectures
that will optimize for not only  the more commonly used metrics such as
 accurate and hardware-efficient, but also the fairness.

Without loss of generality, we choose a widely used dermatological dataset,
the International Skin Imaging Collaboration (ISIC)~\cite{isic2019}, to illustrate
some of the key points of our proposed FL-NAS design. The ISIC dataset
is a large and expanding open-source archive of digital skin images of melanoma, 
one of the most lethal of all skin cancers. The 2019 version of the ISIC dataset
contains 25,331  images of skin lesions, encompassing eight dermatological diseases.
The accuracy of DNNs will be measured by the classification accuracy of these eight skin disease types. The hardware-efficiency will be measured by the DNN model's parameter sizes and 
memory footprint for deployment. To measure fairness across demographic groups,
we segment the dataset based on the images' metadata about the biological genders and ages. We consider the male and female as a biological gender group while dividing the age groups into the young age
group for those under 30 years of old, the middle age group for those between 30 and 65,
and the old age group for those above 65. Below we discuss the fairness metrics for DNNs across
different demographic groups.

\subsection{Fairness Metrics for DNNs}
\noindent
Loosely speaking, fairness can be understood as the requirement that a DNN model's prediction
for populations over different demographic groups should be close to each other.

One quantified fairness metric as proposed by the authors of~\cite{sheng2022larger} is
called \textit{unfairness} scores, that measures the 
average differences of each demographic group's model prediction
accuracy and the overall model prediction accuracy, i.e.,
\begin{equation}
    \text{Unfairness} = \frac{1}{P} \sum_{j=1}^{P} | \text{Acc}_{j} - \text{Acc}_{\text{overall}} |\nonumber
\end{equation}
\noindent
where \( P \) is the total number of demographic groups, \( \text{Acc}_{j} \) 
is the accuracy of the \(j\)-th demographic group, \( \text{Acc}_{\text{overall}} \) is the overall accuracy across all demographic groups.
In other words, the unfairness score gauges the bias across all demographic groups in a dataset,
and the lower the unfairness score, the fairer (and the better) the DNN model.

Another way to quantify the fairness metric as proposed by the authors of~\cite{hardt2016equality}
is to consider the 
opportunity equality for each demographic type within a demographic group, and
it concerns on positive and negative samples.
Accordingly, there are three different fairness metrics: equalized odds (EODD), 
equal opportunity concerning the positive class (EOPP1), and equal 
opportunity concerning the negative class (EOPP2), respectively.

\textbf{Equalized Odds (EODD)} asserts that, given the same opportunity, the True Positive Rate (TPR) and False Positive Rate (FPR) should be consistent across all demographic groups. This measure aims to ensure fairness for both favorable and unfavorable outcomes.
\begin{equation}
    \text{EODD} = \frac{1}{P} \sum_{p=1}^{P} \max \left( \left| \frac{\text{TP}^{(1)}_{p}}{\text{P}^{(1)}_{p}} - \frac{\text{TP}^{(2)}_{p}}{\text{P}^{(2)}_{p}} \right|, \left| \frac{\text{FP}^{(1)}_{p}}{\text{N}^{(1)}_{p}} - \frac{\text{FP}^{(2)}_{p}}{\text{N}^{(2)}_{p}} \right| \right),\nonumber
\end{equation}
\noindent
where \( P \) is the total number of demographic group pairs.
Within each group pair \( p \), 
\( \text{P}_{p}^{(1)} \) and \( \text{P}_{p}^{(2)} \) are the total number of positive instances for the first and second groups, respectively;
\( \text{TP}_{p}^{(1)} \) and \( \text{TP}_{p}^{(2)} \) are the number of true positives for the first and second groups, respectively.
Similarly, \( \text{N}_{p}^{(1)} \) and \( \text{N}_{p}^{(2)} \) are the total number of negative instances for the first  and second groups, respectively;
while \( \text{FP}_{p}^{(1)} \) and \( \text{FP}_{p}^{(2)} \) are the number of false positives for the first and second groups, respectively.

Different from EODD, \textbf{Equal Opportunity with respect to the positive class (EOPP1)} solely focuses on fairness in favorable outcomes. It considers a DNN model's 
outcome as fair when the true positive rate is consistent among all demographic groups, i.e., ensuring that each group has an equal opportunity to receive positive outcomes as follows:
\begin{equation}
    \text{EOPP1} = \frac{1}{P} \sum_{p=1}^{P} \left| \frac{\text{TP}^{(1)}_{p}}{\text{P}^{(1)}_{p}} - \frac{\text{TP}^{(2)}_{p}}{\text{P}^{(2)}_{p}} \right|.\nonumber
\end{equation}

Similarly, \textbf{Equal Opportunity with respect to the negative class (EOPP2)} emphasizes fairness in unfavorable outcomes. It deems a DNN model's outcome fair when the true negative rate is consistent among all demographic groups to ensure fairness in avoiding negative outcomes, i.e.,

\begin{equation}
    \text{EOPP2} = \frac{1}{P} \sum_{p=1}^{P} \left| \frac{\text{TN}^{(1)}_{p}}{\text{N}^{(1)}_{p}} - \frac{\text{TN}^{(2)}_{p}}{\text{N}^{(2)}_{p}} \right|,\nonumber
\end{equation}
\noindent
where, within each group pair \( p \), 
\( \text{TN}_{p}^{(1)} \) and \( \text{TN}_{p}^{(2)} \) are the total number of negative instances for the first and second groups, respectively;
while \( \text{FN}_{p}^{(1)} \) and \( \text{FN}_{p}^{(2)} \) are the number of false negative positives for the first and second groups, respectively.

\subsection{FL-NAS Framework Design}
\noindent
We show the design of FL-NAS framework in Algorithm ~\ref{alg:FLNAS}. The framework takes as input: (1) a proper DNN design template (\textit{Template}), (2) the changeable
parameter ranges for the template, i.e., architecture search space (\textit{Choices}), 
(3) the target hardware deployment environment (\textit{Env}), and (4) the maximum allowable
search iterations (\textit{IterMax}). The output of the framework is
the final DNN architecture (\textit{OptDNN}) with the best evaluation metrics (\textit{OptMetrics}).
The framework consists of three major components: the LLM-based prompt generator (\textit{PromptGenerator}),
the LLM-based design generator (\textit{LLMDesigner}), and the DNN evaluator (\textit{Evaluator}). 

We first initialize two dictionary structures, \textit{archDNNs} and \textit{metricDNNs}, to
store all the searched DNN architectures and their corresponding evaluation metrics, respectively.
We then perform a NAS-like search for newer and better DNN architectures through LLMs.
During each NAS search iteration, we first obtain the best-so-far DNN architecture (\textit{optDNN})
and its corresponding design metrics (\textit{optMetrics});
we then generate  proper prompts (\textit{Prompts}) by calling
the function of \textit{PromptGenerator};
the prompts will be fed into an LLM engine of choice to obtain a newly recommended DNN architecture
(\textit{newDNN}) by calling the function of \textit{LLMDesigner};
and the newly found DNN architecture will then be evaluated for the various design
metrics, all of which will be added to the two dictionary structures, \textit{archDNNs} and \textit{metricDNNs}, for both-keeping purpose.
When the NAS search iteration is exhausted, we will return the final best DNN architecture (\textit{optDNN}) and its corresponding design metrics (\textit{optMetrics}).
Below we discuss each of the three main components in detail.

\begin{algorithm}[t!]
    \footnotesize
    \caption{FL-NAS Framework}\label{alg:FLNAS}
    \begin{algorithmic}[1]
    
    \State \textbf{Input:} \textit{Template}, \textit{Choices}, \textit{Env}, \textit{IterMax}
    \State \textbf{Output:} \textit{OptDNN}, \textit{OptMetrics}
    
    \State Initialize \textit{archDNNs} and \textit{metricDNNs}
    \For{$i \gets 1$ \textbf{to} \textit{IterMax}}
        \State \textit{optDNN}, \textit{optMetrics} $\gets$ \textit{GetBestMetrics}(\textit{archDNNs}, \textit{metricDNNs})
        \State \textit{Prompts} $\gets$ \textit{PromptGenerator}(\textit{optDNN}, \textit{optMetrics}, \textit{Template}, 
        \State \hspace{0.5cm} \textit{Choices}, \textit{Env})
        \State \textit{newDNN} $\gets$ \textit{LLMDesigner}(\textit{Prompts})
        \State \textit{newMetrics} $\gets$ \textit{Evaluator}(\textit{newDNN})
        \State Add \textit{newDNNs} and \textit{optDNNS} to \textit{archDNNs} and \textit{metricDNNs}
    \EndFor    
    \State \Return \textit{GetBestMetrics}(\textit{archDNNs}, \textit{metricDNNs})
    \end{algorithmic}
\end{algorithm}

\subsection{LLM-based Prompt Generator}

\noindent
A key to use LLMs properly for NAS is to design meaningful prompts so that we can
leverage LLMs' strength in understanding the large input context and 
background deep domain knowledge. To do so, we include the detailed
design of the best-so-far DNN architecture along with its design metrics
as part of the prompts. In doing so, we can provide a much richer
input context to the LLM so that it can generate a better choice
for the DNN template's architectural parameters. 

\begin{algorithm}
    \caption{PromptGenerator Function}
    \footnotesize
    \label{alg:prompt-generator}
    \begin{changemargin}{0.3cm}{0cm}
    \begin{flushleft}
    \textbf{Input:} \textit{optDNN}, \textit{optMetrics}, \textit{Template}, \textit{Choices}, \textit{Env} \newline
    \textbf{Output:} \textit{Prompts} \newline    
    \textit{PromptApp} = ``You are an expert in the field of neural architecture search.'' \newline
    \textit{PromptTask} = ``You need to follow the format in \textit{Template} to generate a DNN architecture'' \newline
    \textit{PromptConst} = ``Your task is to design a DNN architecture to process the image from ISIC dataset for image classification task. The current best-performing DNN architecture is \textit{optDNN} and its evaluation is \textit{optMetrics}. 
    
    Your designed DNN will be deployed into \textit{Env}, so you need to make sure you consider the limitations in this running environment.
    
    In evaluation, the metrics are accuracy, fairness, and hardware efficiency. You need to Improve the fairness without decreasing the accuracy. Also notice the accuracy in each demographic group and the overall accuracy. Try to design a DNN to improve the lowest accuracy in certain demographic groups.
    
    Regarding the hardware efficiency metrics, which are number of parameters, latency, and memory, you should try to minimize them so the DNN can run with less resources. But the priority of optimizing hardware efficiency is lower than maintaining high accuracy, which is also lower than improving fairness.
    
    You can insert or remove convolutional layers, use different normalization methods, regularization methods, and try different sizes of kernel. The available range is \textit{Choices}.''
    
    return \textit{PromptApp} + \textit{PromptTask} + \textit{PromptConst} 
    \end{flushleft} 
    \end{changemargin}
\end{algorithm}
Algorithm ~\ref{alg:prompt-generator} shows our detailed prompt designs for LLMs. 
The first part of the prompt (\textit{PromptApp}) hints to the LLM that the application domain
of interest is NAS. The second part of the prompt (\textit{PromptTask}) tells the LLM to
follow the given DNN template for its generation task. The third part of the prompt
(\textit{PromptConst}) details all kinds of design constraints that the LLM needs
to consider in performing its generative task, and this is where the LLM shines over the
traditional EA-based or RL-based algorithms. Instead of
relying on GPUs to conduct costly evaluations of many potential DNN candidates, 
the LLM will digest the given best-so-far DNN design, and its metrics, use its deep
background domain knowledge in terms of hardware deployment, computation efficiency,
fairness, and their relative importance, all being understood in a semantic sense (as compared
to being ``calculated'' explicitly). This is why the LLM-based NAS will be
orders-of-magnitude faster than the traditional NAS, and our experimental results in Section \ref{sec_exp} will also confirm this.

Instead of feeding only one design metric (such as accuracy) to the LLM, we
propose to give a much more detailed description of all design metrics to the LLM. 
We hypothesize that this will provide a richer context for the LLM to
reason about the comparison of those metrics in its internal generation progress,
though we still know little about how the LLM actually works internally. 
An example of our detailed design metrics are as follows: \textit{optMetrics}=``Train Loss: $0.8746$, Train Acc: $58.06\%$, Valid Loss: $0.6438$, Valid Acc: $62.80\%$, Unfairness Score: $1.2983$, EODD: $0.1229$, EOPP1: $0.1224$, EOPP2: $0.0105$, Fairness Detail: [male: ($63.23\%$, $95$), female: ($62.37\%$, $93$), young: ($25.99\%$, $23$), middle\_age: ($59.30\%$, $104$), old: ($81.82\%$, $62$)] Latency: $0.002153$ seconds per image, Throughput: $464.38$ images per second, Peak GPU Memory Usage: $232.88$ MB.''

\subsection{LLM-based Design Generator}
\begin{figure}[t!]
    \centering
    \includegraphics[width=\columnwidth]{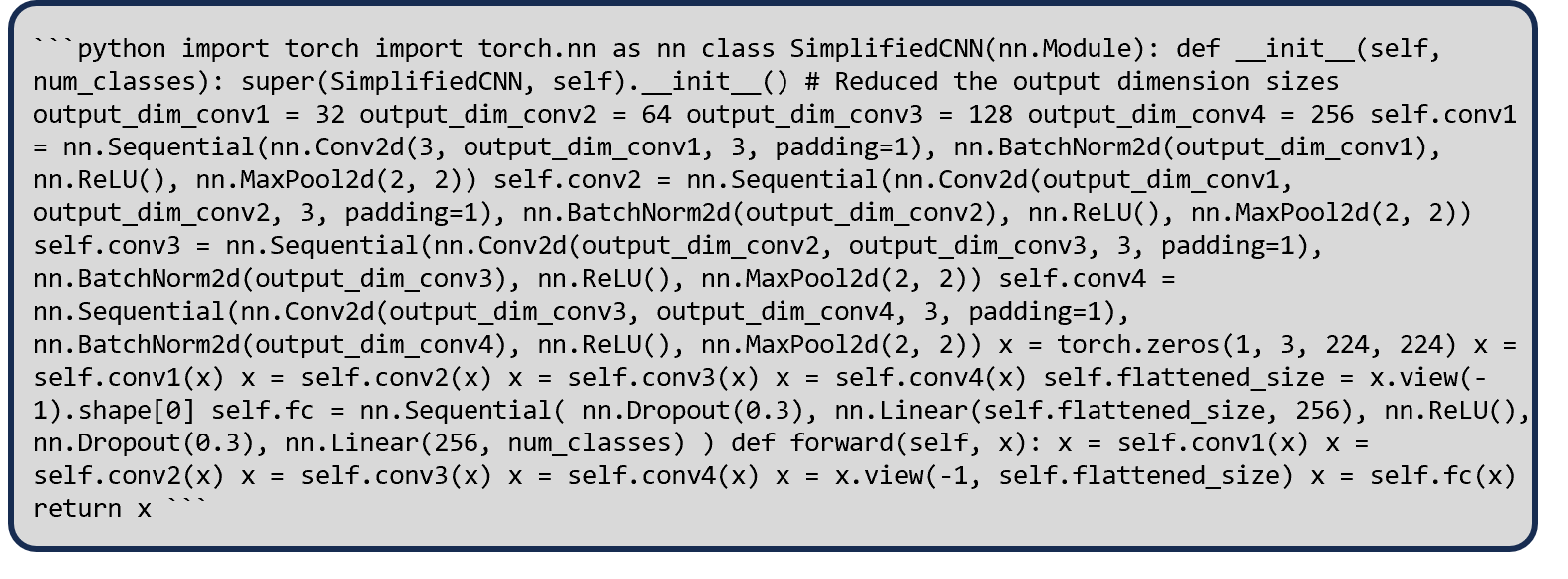}
    \caption{An example of LLM outputs related to DNN.}
    \label{fig:example}
    \vspace{-0.3cm}
  \end{figure}

\noindent
The LLM design generator (\textit{PromptGenerator}) will take the prompts and ask the LLM to generate the outputs.
From the outputs, we will extract the text related to the DNN architecture designs,
and one such example is shown in Figure~\ref{fig:example}. From their, we will format the text into
the proper Python codes for a given deep learning framework (PyTorch in this case),
and check if the extracted Python codes indeed describe a valid DNN architecture. 
If the answer is no, we will continue to ask the LLM to redo the generation task until a valid DNN architecture is obtained. We will further assemble such a valid DNN architecture into
an executable DNN model.

\subsection{DNN Evaluator}
\noindent
The DNN evaluator (\textit{Evaluator})  evaluates the quality of a DNN architecture in 
terms of all considered metrics. It will first train the DNN under the same hyperparameter settings,
and stop the training when the validation loss no longer decreases.
The final train accuracy, train loss, validation accuracy, and validation loss will be recorded.
Moreover, the accuracy of each demographic group will also be calculated as well as the EODD, EOPP1, and EOPP2 metrics. For hardware efficiency, we further record the DNN model's inference latency, throughput, parameter sizes, and GPU memory required to run the model. All of these metrics will be saved as part of the evaluation of the DNN.






\section{Experimental Results}
\label{sec_exp}

\captionsetup[table]{skip=4pt}
\newcolumntype{P}[1]{>{\centering\arraybackslash}p{#1}}

\begin{table*}[t]
\centering
\caption{Comparison of top three FL-NAS models with five baseline DNN models}
\label{tab:evaluation}
\begin{tabular}{c|c|P{1.1cm}P{1.1cm}P{1.1cm}P{1.1cm}P{1.1cm}|P{1.3cm}P{1.4cm}|P{1.9cm}P{1.4cm}}
\toprule
\midrule
\multicolumn{2}{c|}{\multirow{2}{*}{DNN}} & \multicolumn{5}{c|}{\textbf{DNN Evaluation}} & \multicolumn{2}{c|}{\textbf{Hardware Efficiency}} & \multicolumn{2}{c}{\textbf{Latency (s)}} \\
\multicolumn{2}{c|}{} & Accuracy & Unfairness & EODD & EOPP1 & EOPP2 & \# of Para. & Mem (MB) & NVIDIA A10 &  Pi 4 \\
\midrule
\multirow{5}{*}{\rotatebox{90}{Baseline}} 
& ResNet-18 & 66.67\% & 0.3200 & 0.1077 & 0.0726 & 0.1006 & 11,169,858 & 2113.91 & 0.00763 & 5.64 \\
& ResNet-34 & 63.80\% & 0.4354 & 0.1556 & 0.0856 & 0.1335 & 21,278,018 & 3503.10 & 0.01450 & 11.26 \\
& MobileNetV1 & 67.73\% & 0.3498 & 0.1470 & 0.0482 & 0.1287 & 3,199,106 & 1108.89 & 0.00594 & 1.31 \\
& MobileNetV2 & 66.93\% & 0.5778 & 0.1697 & 0.0445 & 0.1620 & 7,054,850 & 2444.69 & 0.01563 & 1.62 \\
& FaHaNa-Fair & 71.73\% & 0.2570 & 0.1247 & 0.0583 & 0.1155 & 5,532,866 & 12864.62 & 0.00644 & 3.42 \\
\midrule
\multirow{3}{*}{\rotatebox{90}{FL-NAS}} 
& Search-1 & \textuparrow 75.20\% & \textdownarrow 0.1044 & 0.1711 & \textdownarrow 0.0346 & 0.1711 & \textdownarrow 3,309,570 & \textdownarrow 211.79 & \textdownarrow 0.00273 & \textdownarrow 0.11 \\
& \textbf{Search-2} & \textuparrow \textbf{73.20\%} & \textdownarrow \textbf{0.0306} & \textdownarrow \textbf{0.0995} & \textdownarrow \textbf{0.0340} & \textdownarrow \textbf{0.0995} & \textdownarrow \textbf{293,954} & \textdownarrow \textbf{232.88} & \textdownarrow \textbf{0.0011} & \textdownarrow \textbf{0.15} \\
& Search-3 & \textuparrow 72.00\% & \textdownarrow 0.1136 & 0.1107 & \textdownarrow 0.0204 & 0.1107 & \textdownarrow 148,098 & \textdownarrow 158.25 & \textdownarrow 0.00241 & \textdownarrow 0.08 \\
\midrule
\bottomrule
\end{tabular}
\end{table*}

\subsection{Experimental Setup}

\noindent
We design our experiments to demonstrate the performance of FL-NAS for solving image classification tasks. The FL-NAS uses the basic convolutional neural network (CNN) design as a \textit{Template}, and the LLM used is the latest ChatGPT-4 engine.

We employ the dataset from the ISIC challenge~\cite{isic2019} that contains eight 
dermatological diseases to be classified. 
To evaluate the fairness metrics (\textit{Unfairness}, \textit{EODD}, \textit{EOPP1} and \textit{EOPP2} for models' predictions,
we further label the dataset based on the images' metadata about the biological genders and ages. We consider the male and female as a biological gender group while dividing the age groups into
three: the young age group (less than 30 years), the middle age group (between 30 and 65),
and the old age group (greater than 65). To avoid the need for high computing resources for
running the experiments, without loss of generality, we 
randomly select $5,000$ images from the total of $25,331$ images from the original
ISIC dataset as the new dataset for our experiments. 
For the chosen $5,000$ images, we ensure a balanced representation of each demographic group to mitigate the potential bias stemming from the training data. The dataset is partitioned into the
the training set, validation set, and test set with the ratio of 70\%, 20\% and 10\%, respectively.
The validation set is used to guide the training process and set the \textit{IterMax}=10 for
FL-NAS.

The hardware-efficiency are measured by DNN models' parameter sizes,  memory footprint for deployment, and latency.
We use the NVIDIA A10 as the target deployment environment (\textit{Env}) during FL-NAS, but the final obtained DNN models are also evaluated on another edge device, Raspberry Pi 4,
for its latency performance only.

The accuracy of models will be measured by the classification accuracy of these eight skin disease types, i.e., $\text{Accuracy} = \frac{\sum_{i=1}^{8} \text{TP}_i}{N}$, where $\text{TP}_i$ is the
true positives for the $i^{th}$ skin disease, and $N$ is the total images.

For comparison purpose, 
four state-of-the-art (SOTA) CNNs are used as a baseline for their robustness in solving
the image classification problems, and they are
ResNet-18~\cite{he2016deep}, ResNet-34~\cite{he2016deep}, 
MobileNetV1~\cite{howard2017mobilenets}, and MobileNetV2~\cite{sandler2018mobilenetv2}.
Out of the many NAS work, the most relevant NAS work to FL-NAS is FaHaNa~\cite{sheng2022larger}, 
as it also considers the same set of metrics as FL-NAS, i.e., fairness, hardware efficiency,
and the conventional accuracy metrics, except that FaHaNa relies on the traditional reinforcement learning framework while FL-NAS is based on LLMs.

\subsection{Evaluations}

\noindent
We report our experimental results as shown in Table~\ref{tab:evaluation} under the same experimental
setups. To better understand
FL-NAS's capabilities, we have included the top three DNNs as found by FL-NAS in
terms of their \textit{Unfairness} scores and \textit{Accuracy}. We name
them as Search-1, search-2, and Search-3 in Table~\ref{tab:evaluation}. 
We employ the notation of \textbf{up arrow} ($\uparrow$) and \textbf{down arrow} ($\downarrow$) to indicate whether the various metric values of each FL-NAS searched DNN are either
higher or lower than any of the baselines for the same metric category. 

We first report the accuracy results. Though we emphasize the novelty of incorporating fairness
(and hardware efficiency) into FL-NAS, accuracy is still one of the most important 
considerations for designing any DNN model. A perfect fair DNN model with low accurate prediction would not be very
useful
in practice. As shown in Table~\ref{tab:evaluation}, we observe, interestingly, that all FL-NAS searched DNNs 
in fact achieved higher accuracy (up to 11.4\%) compared to the baselines.  This shows that the ChatGPT-4 engine
used in FL-NAS is able to find good DNN architectures with high accuracy. 

We then compare the fairness metrics achieved by the three FL-NAS models with the other five baseline models.
We see that the top three FL-NAS models are almost always better than those baseline models across almost all
fairness metrics, and the improvement ranges from 8$\times$ to 19$\times$. 
For example, looking at the \textit{Unfairness} score
column in Table~\ref{tab:evaluation}, the best FL-NAS achieves a score of 0.0306, while the best and the worst
scores for the baselines are 0.2570 and 0.5778, respectively. Similar improvements are also observed
for the \textit{EODD}, \textit{EOPP1}, and \textit{EOPP2}, fairness metrics. 
This clearly demonstrates the superior capabilities
of LLM in understanding the semantic meaning of fairness, 
hence generating high-quality DNN designs
with improved fairness metrics. 

Finally, we report the hardware efficiency comparison for the various models. We first show the comparison in terms of
model parameter numbers and peak memory footprint. It is clear from Table~\ref{tab:evaluation} that 
the top three FL-NAS models
are all much smaller than the (even the best) baseline models. For example, the parameter numbers of the 
Search-2 model can be orders of magnitude smaller than the best baseline model; while the memory size
reduction can be up to 10$\times$. Next, we show the comparison on inference latency when models are
deployed onto two different hardware platforms, the NVIDIA A10 GPU and the Raspberry Pi 4, respectively.
We show the latency number for a batch size of two so that we can compare the latency numbers for both platforms.
As is clear, the FL-NAS models are significantly faster (more than 6$\times$) than all the baseline models. All of these improvements again demonstrate the LLM's great potential 
in helping FL-NAS to find new DNN models that are computationally efficient as well.

Note that the top three performing FL-NAS models are found through interacting with LLMs up to to 10 times (\textit{IterMax}=3),
which takes about three to five GPU hours on the NVIDIA A10 platform to train the candidate DNN models.
This is in stark contrast to the traditional NAS approaches, where thousands of GPU hours are typical in finding
a reasonably good DNN model. This is another benefit of using an LLM-based NAS framework like FL-NAS.

In summary, our experimental results seem to convincingly show us that our LLM-based FL-NAS framework
is efficient in finding high-performing DNN models that are accurate, fair, and computationally efficient.



\section{Conclusion}
\label{sec_diss}

\noindent
This paper has proposed a novel LLM-based NAS framework, FL-NAS, that can  take into account various
design metrics in a flexible manner in designing a new DNN model.
For the first time, we have demonstrated that FL-NAS can perform efficient NAS search 
while optimizing for accuracy, fairness, and hardware efficiency simultaneously.
Our experimental results have further illustrated the effectiveness of the proposed framework when compared to the
traditionally human-designed DNNs and other NAS approaches under similar design settings. Our preliminary exploration
has  convincingly
shown that the LLM-based NAS approach warrants more in-depth research to fully leverage the power of LLMs.

\bibliographystyle{IEEEtran}
\bibliography{sample-base}

\begin{thebibliography}{10}
\providecommand{\url}[1]{#1}
\csname url@samestyle\endcsname
\providecommand{\newblock}{\relax}
\providecommand{\bibinfo}[2]{#2}
\providecommand{\BIBentrySTDinterwordspacing}{\spaceskip=0pt\relax}
\providecommand{\BIBentryALTinterwordstretchfactor}{4}
\providecommand{\BIBentryALTinterwordspacing}{\spaceskip=\fontdimen2\font plus
\BIBentryALTinterwordstretchfactor\fontdimen3\font minus \fontdimen4\font\relax}
\providecommand{\BIBforeignlanguage}[2]{{%
\expandafter\ifx\csname l@#1\endcsname\relax
\typeout{** WARNING: IEEEtran.bst: No hyphenation pattern has been}%
\typeout{** loaded for the language `#1'. Using the pattern for}%
\typeout{** the default language instead.}%
\else
\language=\csname l@#1\endcsname
\fi
#2}}
\providecommand{\BIBdecl}{\relax}
\BIBdecl

\bibitem{elsken2019neural}
T.~Elsken, J.~H. Metzen, and F.~Hutter, ``Neural architecture search: A survey,'' \emph{The Journal of Machine Learning Research}, vol.~20, no.~1, pp. 1997--2017, 2019.

\bibitem{chu2021fairnas}
X.~Chu, B.~Zhang, and R.~Xu, ``Fairnas: Rethinking evaluation fairness of weight sharing neural architecture search,'' in \emph{Proceedings of the IEEE/CVF International Conference on computer vision}, 2021, pp. 12\,239--12\,248.

\bibitem{liu2021fnas}
J.~Liu, M.~Zhang, Y.~Sun, B.~Liu, G.~Song, Y.~Liu, and H.~Li, ``Fnas: Uncertainty-aware fast neural architecture search,'' \emph{arXiv preprint arXiv:2105.11694}, 2021.

\bibitem{sheng2022larger}
Y.~Sheng, J.~Yang, Y.~Wu, K.~Mao, Y.~Shi, J.~Hu, W.~Jiang, and L.~Yang, ``The larger the fairer? small neural networks can achieve fairness for edge devices,'' in \emph{Proceedings of the 59th ACM/IEEE Design Automation Conference}, 2022, pp. 163--168.

\bibitem{jia2023importance}
Z.~Jia, J.~Chen, X.~Xu, J.~Kheir, J.~Hu, H.~Xiao, S.~Peng, X.~S. Hu, D.~Chen, and Y.~Shi, ``The importance of resource awareness in artificial intelligence for healthcare,'' \emph{Nature Machine Intelligence}, pp. 1--12, 2023.

\bibitem{kaissis2020secure}
G.~A. Kaissis, M.~R. Makowski, D.~R{\"u}ckert, and R.~F. Braren, ``Secure, privacy-preserving and federated machine learning in medical imaging,'' \emph{Nature Machine Intelligence}, vol.~2, no.~6, pp. 305--311, 2020.

\bibitem{han2015deep}
S.~Han, H.~Mao, and W.~J. Dally, ``Deep compression: Compressing deep neural networks with pruning, trained quantization and huffman coding,'' \emph{arXiv preprint arXiv:1510.00149}, 2015.

\bibitem{yan2022reliability}
Z.~Yan, X.~S. Hu, and Y.~Shi, ``On the reliability of computing-in-memory accelerators for deep neural networks,'' in \emph{System Dependability and Analytics: Approaching System Dependability from Data, System and Analytics Perspectives}.\hskip 1em plus 0.5em minus 0.4em\relax Springer, 2022, pp. 167--190.

\bibitem{yan2022swim}
------, ``Swim: Selective write-verify for computing-in-memory neural accelerators,'' in \emph{Proceedings of the 59th ACM/IEEE Design Automation Conference}, 2022, pp. 277--282.

\bibitem{nagel2021white}
M.~Nagel, M.~Fournarakis, R.~A. Amjad, Y.~Bondarenko, M.~Van~Baalen, and T.~Blankevoort, ``A white paper on neural network quantization,'' \emph{arXiv preprint arXiv:2106.08295}, 2021.

\bibitem{OpenAI2023GPT4TR}
\BIBentryALTinterwordspacing
OpenAI, ``Gpt-4 technical report,'' \emph{ArXiv}, vol. abs/2303.08774, 2023. [Online]. Available: \url{https://api.semanticscholar.org/CorpusID:257532815}
\BIBentrySTDinterwordspacing

\bibitem{touvron2023llama}
H.~Touvron, T.~Lavril, G.~Izacard, X.~Martinet, M.-A. Lachaux, T.~Lacroix, B.~Rozi{\`e}re, N.~Goyal, E.~Hambro, F.~Azhar \emph{et~al.}, ``Llama: Open and efficient foundation language models,'' \emph{arXiv preprint arXiv:2302.13971}, 2023.

\bibitem{biderman2023pythia}
S.~Biderman, H.~Schoelkopf, Q.~G. Anthony, H.~Bradley, K.~O’Brien, E.~Hallahan, M.~A. Khan, S.~Purohit, U.~S. Prashanth, E.~Raff \emph{et~al.}, ``Pythia: A suite for analyzing large language models across training and scaling,'' in \emph{International Conference on Machine Learning}.\hskip 1em plus 0.5em minus 0.4em\relax PMLR, 2023, pp. 2397--2430.

\bibitem{anil2023palm}
R.~Anil, A.~M. Dai, O.~Firat, M.~Johnson, D.~Lepikhin, A.~Passos, S.~Shakeri, E.~Taropa, P.~Bailey, Z.~Chen \emph{et~al.}, ``Palm 2 technical report,'' \emph{arXiv preprint arXiv:2305.10403}, 2023.

\bibitem{yan2023viability}
Z.~Yan, Y.~Qin, X.~S. Hu, and Y.~Shi, ``On the viability of using {LLMs} for {SW/HW} co-design: An example in designing {CiM} {DNN} accelerators,'' \emph{arXiv preprint arXiv:2306.06923}, 2023.

\bibitem{zheng2023can}
M.~Zheng, X.~Su, S.~You, F.~Wang, C.~Qian, C.~Xu, and S.~Albanie, ``Can gpt-4 perform neural architecture search?'' \emph{arXiv preprint arXiv:2304.10970}, 2023.

\bibitem{wang2023graph}
H.~Wang, Y.~Gao, X.~Zheng, P.~Zhang, H.~Chen, and J.~Bu, ``Graph neural architecture search with gpt-4,'' \emph{arXiv preprint arXiv:2310.01436}, 2023.

\bibitem{li2023diffnas}
W.~Li, X.~Su, S.~You, F.~Wang, C.~Qian, and C.~Xu, ``Diffnas: Bootstrapping diffusion models by prompting for better architectures,'' \emph{arXiv preprint arXiv:2310.04750}, 2023.

\bibitem{zoph2016neural}
B.~Zoph and Q.~V. Le, ``Neural architecture search with reinforcement learning,'' \emph{arXiv preprint arXiv:1611.01578}, 2016.

\bibitem{liu2021survey}
Y.~Liu, Y.~Sun, B.~Xue, M.~Zhang, G.~G. Yen, and K.~C. Tan, ``A survey on evolutionary neural architecture search,'' \emph{IEEE transactions on neural networks and learning systems}, 2021.

\bibitem{baker2017accelerating}
B.~Baker, O.~Gupta, R.~Raskar, and N.~Naik, ``Accelerating neural architecture search using performance prediction,'' \emph{arXiv preprint arXiv:1705.10823}, 2017.

\bibitem{zhang2019neural}
X.~Zhang, W.~Jiang, Y.~Shi, and J.~Hu, ``When neural architecture search meets hardware implementation: from hardware awareness to co-design,'' in \emph{2019 IEEE Computer Society Annual Symposium on VLSI (ISVLSI)}.\hskip 1em plus 0.5em minus 0.4em\relax IEEE, 2019, pp. 25--30.

\bibitem{stamoulis2019single}
D.~Stamoulis, R.~Ding, D.~Wang, D.~Lymberopoulos, B.~Priyantha, J.~Liu, and D.~Marculescu, ``Single-path nas: Designing hardware-efficient convnets in less than 4 hours,'' in \emph{Joint European Conference on Machine Learning and Knowledge Discovery in Databases}.\hskip 1em plus 0.5em minus 0.4em\relax Springer, 2019, pp. 481--497.

\bibitem{li2020edd}
Y.~Li, C.~Hao, X.~Zhang, X.~Liu, Y.~Chen, J.~Xiong, W.-m. Hwu, and D.~Chen, ``{EDD}: Efficient differentiable {DNN} architecture and implementation co-search for embedded ai solutions,'' in \emph{2020 57th ACM/IEEE Design Automation Conference (DAC)}.\hskip 1em plus 0.5em minus 0.4em\relax IEEE, 2020, pp. 1--6.

\bibitem{cai2018proxylessnas}
H.~Cai, L.~Zhu, and S.~Han, ``Proxylessnas: Direct neural architecture search on target task and hardware,'' \emph{arXiv preprint arXiv:1812.00332}, 2018.

\bibitem{zhou2020econas}
D.~Zhou, X.~Zhou, W.~Zhang, C.~C. Loy, S.~Yi, X.~Zhang, and W.~Ouyang, ``Econas: Finding proxies for economical neural architecture search,'' in \emph{Proceedings of the IEEE/CVF Conference on computer vision and pattern recognition}, 2020, pp. 11\,396--11\,404.

\bibitem{Li_2023_ICCV}
Y.~Li, J.~Li, C.~Hao, P.~Li, J.~Xiong, and D.~Chen, ``Extensible and efficient proxy for neural architecture search,'' in \emph{Proceedings of the IEEE/CVF International Conference on Computer Vision (ICCV)}, October 2023, pp. 6199--6210.

\bibitem{isic2019}
\BIBentryALTinterwordspacing
(2019) Skin lesion analysis. [Online]. Available: \url{https://challenge2019.isic-archive.com/}
\BIBentrySTDinterwordspacing

\bibitem{hardt2016equality}
M.~Hardt, E.~Price, and N.~Srebro, ``Equality of opportunity in supervised learning,'' \emph{Advances in neural information processing systems}, vol.~29, 2016.

\bibitem{he2016deep}
K.~He, X.~Zhang, S.~Ren, and J.~Sun, ``Deep residual learning for image recognition,'' in \emph{Proceedings of the IEEE conference on computer vision and pattern recognition}, 2016, pp. 770--778.

\bibitem{howard2017mobilenets}
A.~G. Howard, M.~Zhu, B.~Chen, D.~Kalenichenko, W.~Wang, T.~Weyand, M.~Andreetto, and H.~Adam, ``Mobilenets: Efficient convolutional neural networks for mobile vision applications,'' \emph{arXiv preprint arXiv:1704.04861}, 2017.

\bibitem{sandler2018mobilenetv2}
M.~Sandler, A.~Howard, M.~Zhu, A.~Zhmoginov, and L.-C. Chen, ``Mobilenetv2: Inverted residuals and linear bottlenecks,'' in \emph{Proceedings of the IEEE conference on computer vision and pattern recognition}, 2018, pp. 4510--4520.

\end{thebibliography}

\end{document}